# Toward Case-Based Preference Elicitation: Similarity Measures on Preference Structures

Vu Ha    Peter Haddawy
Decision Systems and Artificial Intelligence Lab
Dept. of EE&CS
University of Wisconsin-Milwaukee
Milwaukee, WI 53201
{vu,haddawy}@cs.uwm.edu

## Abstract

While decision theory provides an appealing normative framework for representing rich preference structures, eliciting utility or value functions typically incurs a large cost. For many applications involving interactive systems this overhead precludes the use of formal decision-theoretic models of preference. Instead of performing elicitation in a vacuum, it would be useful if we could augment directly elicited preferences with some appropriate default information. In this paper we propose a case-based approach to alleviating the preference elicitation bottleneck. Assuming the existence of a population of users from whom we have elicited complete or incomplete preference structures, we propose eliciting the preferences of a new user interactively and incrementally, using the closest existing preference structures as potential defaults. Since a notion of closeness demands a measure of distance among preference structures, this paper takes the first step of studying various distance measures over fully and partially specified preference structures. We explore the use of Euclidean distance, Spearman's footrule, and define a new measure, the probabilistic distance. We provide computational techniques for all three measures.

## 1 INTRODUCTION

We are interested in the problem of building interactive systems for which task objectives will be communicated in the form of user preferences. Utility theory, the branch of decision theory that deals with representation of preferences, provides a rich normative framework that is capable of capturing such aspects of preferences as tradeoffs among objectives and attitudes toward risk. But the task of eliciting a utility function is typically time consuming and tedious. For many applications involving interactive systems such overhead can preclude the use of utility theory since the cost in time and effort may be too large relative to the value of the sought after solution or because time may simply be limited. For example, we would be very unhappy if a system designed to help us select a video to watch required as much time as the entire length of the film just to elicit our preferences.

To reduce elicitation overhead, practitioners typically make use of assumptions (e.g. additive independence) that simplify the elicitation task by allowing a high-dimensional utility function to be decomposed into a simple combination of lower-dimensional sub-utility functions. But even under such assumptions, elicitation of a complete utility function can still be too time consuming and, furthermore, the assumptions preclude representation of many kinds of interesting and common preferences.

In previous work [6], we investigated an approach in which we first elicit partial preference information to produce a set of candidate solutions. We then use the set of candidate solutions to identify the additional information to elicit that would likely be most helpful in narrowing down this set. But in order to be able to make useful inferences based only on some preference information, we were forced to assume additive independence of the underlying utility function and to assume that the sub-utility functions are known.

In contrast, Linden, etal. [11] supplement elicited preferences with default information to obtain a complete utility function, which is to be continually adjusted based on the user's feedback. This default preference information represents preferences that are assumed applicable to all users, such as preferences for reduced cost. The user is presented with the optimal solution according to the constructed utility function, together with some extreme points in the solution space (e.g., cheapest and shortest-time flights). The utility function is then modified based on the user's critiques. This approach too assumes an additive utility model in order to facilitate the incorporation of user feedback.

In this paper we also take the approach of supplementing elicited preferences with default information. But rather than applying one default uniformly to all users, we use elicited preferences to select an appropriate default from a set of defaults. Specifically, we investigate a case-based approach to providing such default information. The idea is based on the observation that people tend to form clusters according to their preferences or tastes, an observation that has been analyzed in a large amount of literature in the area of market segmentation [5]. We envision our system to maintain a population of users with their preferences partially or completely specified in a given domain. When encountering a new user $A$, the system elicits some preference information from $A$ and then determines which user in the population has the preference structure that is closest to the preference structure of $A$. The preference structure of that user will be used to determine an initial default representation of $A$'s preferences. In contrast with the previously discussed work of [6] and [11], we do not make any restrictive assumptions concerning the form of the underlying utility functions.

Realization of this approach to elicitation requires that we have a distance measure on preference structures. In this paper we investigate various distance measures from both theoretical and computational perspectives. The rest of the paper is organized as follows. Section 2 deals with distance measures among *completely specified preference structures*. We revisit some well-known measures such as *Spearman's footrule* and *Euclidean distance* and propose to extend these measures to be defined among utility functions. We also propose to define the *probabilistic distance* that measures the difference between two preference structures by the probability that they disagree on the rankings of two random alternatives. In Section 3, we extend all of these definitions to define distance measures among *partially specified preference structures*. We provide a computational technique for approximating these distance measures. In Section 4, we describe MovieFinder, an experimental recommender system that recommends movies for people based on this case-based preference elicitation approach. We discuss related work and future research in Section 5.

## 2 DISTANCE MEASURES ON PREFERENCE ORDERS

We start out with a brief review of utility theory for decision making. The reader is refered to [10] for more details.

The process of making decisions is generally modeled as the identification of the optimal alternative(s) from a set $M$ of alternatives, using in effect a weak order $\prec$, i.e. an asymmetric ($a \prec b \Rightarrow b \not\prec a$), negatively transitive ($a \not\prec b, b \not\prec c \Rightarrow a \not\prec c$) binary relation on the set of alternatives. We will call this relation the *preference structure* of the decision maker: $a \prec b$ indicates that the decision maker *prefers* alternative $b$ to alternative $a$. When neither of the two alternatives is prefered to the other ($a \not\prec b, b \not\prec a$), we say that the decision maker is *indifferent* between them and denote this relation by $a \sim b$. We also use the notation $a \preceq b$ to denote that the decision maker prefers $b$ to $a$ or is indifferent between them. If for all distinct alternatives $a$ and $b$, either $a \prec b$ or $b \prec a$, the $\prec$ relation is said to be a *strict order*.

An important technique that is often used in association with preference orders is the use of *consistent functions* that capture preference orders.

**Definition 1** *A real-valued function $f : M \to \Re$ is said to be* consistent *with a preference order $\prec$ on $M$ if for all $a, b \in M$, $a \prec b \Leftrightarrow f(a) < f(b)$.*

### 2.1 THE CERTAINTY CASE

When a decision problem involves no uncertainty about the alternatives, we call the alternatives *outcomes*, and denote the set of outcomes by $\Omega$. For ease of exposition, we will assume throughout the paper that $\Omega$ is finite and $\Omega = \{s_1, \ldots, s_n\}$. It can be proven [10] that for any preference order $\prec$ over $\Omega$ there exists a function $v$, called a *value function*, that is consistent with $\prec$, and we sometimes write $\prec$ as $\prec_v$ to emphasize this relationship. In the case when $\prec$ is a strict order, and suppose that the elements of $\Omega$ are indexed in such a way that $s_1 \prec s_2 \prec \ldots \prec s_n$, we can define a value function $v$ as $v(s_j) = j, j = 1, \ldots, n$ and call this the *height* of $s_j$. Two value functions that induce identical orders are said to be *strategically equivalent*. Otherwise, they are said to be *strategically different*.

Suppose that there are two users with corresponding preference orders $\prec_1$ and $\prec_2$, which are weak orders on the finite set $\Omega = \{s_1, s_2, \ldots, s_n\}$ of outcomes. We study three distance measures between these two preference orders: Spearman's footrule, Euclidean distance, and probabilistic distance.

On the set of strict orders, the classical distance measure is Spearman's footrule [16]. Suppose that $h_{ij}, i = 1, 2, j = 1, \ldots, n$ is the height of $s_j$ with respect to $\prec_i$. Then the Spearman's footrule is defined as:

**Spearman's footrule** : $\delta_S(\prec_1, \prec_2) := \frac{1}{2} \sum_{i=1}^{n} |h_{1j} - h_{2j}|$.

It is well-known that Spearman's footrule is a metric on strict orders and has the range $[0, \lfloor n^2/4 \rfloor]$ (see, for example [4]). The three requirements for a measure to be a metric are the following (for all strict orders $\prec_i, i = 1, 2, 3$):

(i) **Reflexivity.** $d(\prec_1, \prec_2) \geq 0$,

"=" iff $\prec_1$ and $\prec_2$ are identical.

(ii) **Symmetry.** $d(\prec_1, \prec_2) = d(\prec_2, \prec_1)$.

(iii) **Triangle Inequality.** $d(\prec_1, \prec_3) \leq d(\prec_1, \prec_2) + d(\prec_2, \prec_3)$.

Another popular distance measure among strict orders is the Euclidean distance, defined as:

**Euclidean distance :** $\delta_E(\prec_1, \prec_2) = \sqrt{\sum_{i=1}^{n}(h_{1j} - h_{2j})^2}$.

The Euclidean distance and Spearman's footrule are defined on the set of strict orders and it is not obvious how to extend their definitions to weak orders. This is a limitation, since we do not want to rule out cases when there are equally prefered alternatives. In this paper we investigate yet another distance measure, called *probabilistic distance* that is defined on the broader class of weak orders. This distance measure captures the intuition that the difference in preferences of two users should be proportional to the chance that a uniformly randomly chosen pair $(a, b)$ of alternatives will cause a conflict between the two users, i.e, the two users will rank $a$ and $b$ differently. We use an indicator function to capture conflicts; a conflict occurs when the indicator function takes on the value 1:

$$c_{\prec_1, \prec_2}(a, b) = \begin{cases} 1 & \text{if } (a \preceq_1 b \wedge b \prec_2 a) \vee \\ & (a \prec_1 b \wedge b \preceq_2 a) \vee \\ & (a \preceq_2 b \wedge b \prec_1 a) \vee \\ & (a \prec_2 b \wedge b \preceq_1 a) \\ 0 & \text{otherwise.} \end{cases}$$

Given this definition of conflict, the distance between two weak orders $(\Omega, \prec_1)$ and $(\Omega, \prec_2)$ is defined as:

$$\delta_P(\prec_1, \prec_2) = \Pr(\prec_1 \& \prec_2 \text{ rank } s_j \text{ and } s_k \text{ differently})$$
$$= \frac{2}{n(n-1)} \sum_{1 \leq j < k \leq n} c_{\prec_1, \prec_2}(s_j, s_k).$$

where the last equality means that the probability of conflict is the proportion of conflicting pairs of outcomes (out of $n(n-1)/2$ pairs).

**Proposition 1** *The probabilistic distance on the set of weak orders on $\Omega$ is a metric with range $[0, 1]$.*

*Proof:* It is evident that the probabilistic distance only takes values between 0 and 1, the distance between two identical orders is zero, and zero distance implies two identical weak orders. The symmetry of the distance function trivially follows from the symmetry of the conflict function. Finally, to prove the triangle inequality, we note that for all weak orders $\prec_i, i = 1, 2, 3$ and alternatives $a, b$, $c_{\prec_1, \prec_3}(a, b) = 1$ implies either $c_{\prec_1, \prec_2}(a, b) = 1$ or $c_{\prec_2, \prec_3}(a, b) = 1$, and for all events $X, Y$, $\Pr(X \vee Y) \leq \Pr(X) + \Pr(Y)$. □

**Example 1** *Suppose that there are three kinds of night entertainment in Milwaukee: going to watch a basketball game (B), a movie (M), and going to a blues pub (P). Suppose that Xaviera's preference order $\prec_X$ is such that $B \prec_X M \prec_X P$ and Yvette's preference order $\prec_Y$ is such that $M \prec_Y P \prec_Y B$. The distance between their preferences according to Spearman's footrule, Euclidean distance, and probabilistic distance, is $\delta_S(\prec_X, \prec_Y) = 2$, $\delta_E(\prec_X, \prec_Y) = \sqrt{6}$, $\delta_P(\prec_X, \prec_Y) = 2/3$, respectively. If we want these distances to be normalized so that their ranges are the interval $[0, 1]$, by dividing by their correponding upper bounds, then the normalized distances are 1, $\sqrt{3/4} = .8660$, and $2/3 = .6667$, respectively.*

*Now suppose that Zelda's preference order $\prec_Z$ is such that $P \prec_Z M \prec_Z B$, then the normalized distance between $\prec_X$ and $\prec_Z$, according to Spearman's footrule, Euclidean distance, and probabilistic distance, is 1, 1, and 1 (two orders that are reverses of one another have maximum distance).*

It is clear from Example 1 that according to Euclidean ($\delta_E$) and probabilistic distances ($\delta_P$), the preferences of Yvette are closer to Xaviera's than Zelda's are, while according to Spearman's footrule ($\delta_S$), they are equally distant. We say that two distance measures $\delta_1$ and $\delta_2$ on a set $M$ are *relatively equivalent* if for all $a, b, c \in M$, $\delta_1(a, b) < \delta_1(a, c) \Leftrightarrow \delta_2(a, b) < \delta_2(a, c)$. For example, the Euclidean distance is relatively equivalent to the negative of the rank order correlation coefficient $(-\rho)$. On the other hand, $\delta_S$ is not relatively equivalent to $\delta_E$ and $\delta_P$. It can be shown that $\delta_E$ and $\delta_P$ are also not relatively equivalent.

Intuitively, if two distance measures on $M$ are relatively equivalent, then for any element $a \in M$ and a subset $K \subseteq M$ of $M$, the sets of elements of $K$ that are closest to $a$ according to the two measures are identical. Since $\delta_S$, $\delta_E$, and $\delta_P$ are relatively different, it remains an open question which one we should use for our framework.

## 2.2 THE UNCERTAINTY CASE

When the decision alternatives are uncertain, they are usually modeled by probability distributions over outcomes and are called *prospects*. We denote the set of all probability distributions over $\Omega$ by $\mathcal{S}$. The central result of utility theory is a representation theorem that identifies a set of conditions guaranteeing the existence of a function consistent with the preferences of a decision maker [19, 14]. The theorem states that if the preference order of a decision maker satisfies a few "rational" properties, then there exists a real-valued function, called a *utility function* $u : \Omega \to \Re$, over outcomes such that $p \prec q \Leftrightarrow \langle p, u \rangle < \langle q, u \rangle$. Here $\langle p, u \rangle$, the inner product of the probability vector $p$ and the utility vector $u$, is the expected value of function $u$ with respect to the distribution $p$. It is often convenient to extend $u$, by means of expectation, to a function $u : \mathcal{S} \to \Re$ that maps a prospect $p \in \mathcal{S}$ to $\langle p, u \rangle$. This function is clearly consistent with the preference order $(\mathcal{S}, \prec)$, and we sometimes write $\prec$ as $\prec_u$ to emphasize this relationship.

In this paper, we assume that the preference structures (which are weak orders over the set of prospects) of the users satisfy the required "rational" conditions and thus can be represented using utility functions. We now focus attention on defining a distance measure among such preference structures. We investigate two approaches to defining a distance measure: utility-based and non-utility-based, according to whether utility functions of the preference structures are *explicitly* used in the definitions.

**Utility-Based Distance Measures**

Suppose that the preference structures $(\mathcal{S}, \prec_1)$ and $(\mathcal{S}, \prec_2)$ of two users are represented by their corresponding utility functions $u_1, u_2 : \Omega \to \Re$. We shall define the distance between the preference structures $\prec_1$ and $\prec_2$ as a function of two utility functions $u_1$ and $u_2$, which can be view as two vectors in the $n$-dimension vector space $\Re^n$.

It is well-known that utility functions are unique up to a positive linear transformation, i.e., if $v_1$ is also a utility function that represents $\prec_1$, then there are $\alpha, \beta \in \Re, \alpha > 0$ such that $v_1 = \alpha u_1 + \beta$. $u_1$ and $v_1$ are called *strategically equivalent* utility functions. Thus, a distance measure among preference structures should be defined on the strategic equivalence classes of $\Re^n$. This can be done by selecting a representative vector for each equivalence class, and define some distance measure such as Euclidean, or Spearman's footrule-like distance among the representatives. To select a representative for each equivalence class, we can specify a set of conditions so that exactly one member of each equivalence class satisfies those conditions and thus will be chosen as the representative for that class.

**Example 2** *Suppose that Xaviera and Yvette are involved in a decision problem involving lotteries over three possible outcomes a, b, and c. Furthermore suppose that Xaviera's preferences and attitude toward risks are captured by a utility function $u_X$ such that $u_X(a) = 0, u_X(b) = 1, u_X(c) = 2$, and Yvette's preferences and attitude toward risks are captured by a utility function $u_Y$ such that $u_Y(a) = 1, u_Y(b) = 3, u_Y(c) = 4$. Note that Xaviera and Yvette agree on the ranking of certain outcomes, but have different attitudes toward risky decisions. Suppose that we require that the representatives for strategic equivalence classes must have 0 as their minimum and 1 as their maximum values, i.e., we scale the utility functions from 0 to 1. With respect to these conditions, the representative utility functions of Xaviera and Yvette are $u_X$ and $u_Y$ where $u_X(a) = 0, u_X(b) = 1/2, u_X(c) = 1$, and $u_Y(a) = 0, u_Y(b) = 2/3, u_X(c) = 1$. Thus, the Euclidean ($\delta_E$) and Spearman footrule-like ($\delta_S$) distance between the preference structures of Xaviera and Yvette are: $\delta_E(u_X, u_Y) = \sqrt{\sum_s (u_X(s) - u_Y(s))^2} = 1/6$ and $\delta_S(u_X, u_Y) = 1/2 \sum_s |u_X(s) - u_Y(s)| = 1/12$.*

**Non-Utility-Based Distance Measures**

Note that the above utility-based definitions of distance measures are sensitive to the choice of the conditions to select representative utility functions. Since there appears no rule of thumb to select the representatives, these definitions are rather ad-hoc. The same can be said of the Euclidean distance and Spearman's footrule in the certainty case, since the used heights $\{1, 2, \ldots, n\}$ of the outcomes, which are a particular choice of value functions, are also arbitrary; any increasing sequence of numbers can be used instead.

The probabilistic distance overcomes this tricky problem by relying on the preference structures themselves, rather than a particular choice of value functions. This idea can be extended to the uncertainty case as follows. The distance between two preference orders $(\mathcal{S}, \prec_1)$, $(\mathcal{S}, \prec_2)$ is defined as: [1]

$$\delta_P(\prec_1, \prec_2) = \Pr(\prec_1 \& \prec_2 \text{ rank } p \text{ and } q \text{ differently})$$
$$= \int_{\mathcal{S}} \int_{\mathcal{S}} c_{\prec_1, \prec_2}(p, q) dp dq,$$

where the conflict function $c_{\prec_1, \prec_2}$ is defined over a pair $(p, q)$ of prospects in the obvious way. Here we take

---

[1] Note that although this formal definition of the probabilistic distance does not involve utility functions, any operational definition of this measure inevitably does. In fact, any specification of a preference structure over the (infinite) set of prospects has to involve utility functions.

*all* probability distributions into consideration. In the case when the set of candidates is only a subset of $\mathcal{S}$, then it would be more reasonable to integrate over only that set [2]. Based on this definition, we can define the probabilistic distance of two utility functions $u_1, u_2$ to be the probabilistic distance of the two preference structures they represent: $\delta_P(u_1, u_2) = \delta_P(\prec_{u_1}, \prec_{u_2})$.

**Proposition 2** *The probabilistic distance on the set of weak orders on $\mathcal{S}$ is a metric with range $[0,1]$.*

*Proof:*[Sketch] The proof of this proposition is identical to that of Proposition 1, except for the somewhat non-obvious part that shows that if the distance between two utility functions is 0, then they must be positive linear transformations of each other. The idea here is that for two utility functions $u_1$ and $u_2$, the subset of $\mathcal{S} \times \mathcal{S}$ that consists of pairs of prospects that cause conflict between $\prec_{u_1}$ and $\prec_{u_2}$ is a union of two convex cones, and it can be shown that this union has zero volume if and only if the two utility functions $u_1$ and $u_2$ are positive linear transformations of each other. □

**Example 3 (Continuation of Example 2)** *The probabilistic distance between preference structures of Xaviera ($\prec_X$) and Yvette ($\prec_Y$) is:*

$$d(\prec_X, \prec_Y)$$
$$= \int_\mathcal{S} \int_\mathcal{S} c_{\prec_{u_X}, \prec_{u_Y}}(p,q) dp dq$$
$$= \int_\mathcal{S} \int_\mathcal{S} Neg[(u_X \circ (p-q))(u_Y \circ (p-q))] dp dq$$
$$= 1/9.$$

*Here $p$ and $q$ run over the probability simplex $\mathcal{S}$, which is the equilateral triangle in $R^3$ with vertices $(0,0,1), (0,1,0), (1,0,0)$, $u_X = (0,1,2)$, $u_Y = (0,2,3)$, and $Neg$ is the function that returns 1 if its argument is negative and 0 otherwise.*

*Now suppose that Zelda's utility function is $u_Z = (0,2,1)$, which implies that, unlike Xaviera and Yvette, she prefers $c$ with certainty to $b$ with certainty. Calculations show that $\delta(u_X, u_Z) = 1/3 > 1/9$, which is what we would expect: Yvette's preferences are more similar to Xaviera's than Zelda's are.*

Closely related to the distance concept is the *similarity* concept in fuzzy-set theory [20, 12]. A similarity relation $s$ is a binary fuzzy relation on a set $U$ that satisfies the following three properties, $\forall u,v,w \in U$:

(i) **Reflexivity.** $s(u,u) = 1$.

(ii) **Symmetry.** $s(u,v) = s(v,u)$.

(iii) **∗-Transitivity.** $s(u,v) * s(v,w) \leq s(u,w)$,

---
[2]Provided that the set is measurable.

where $*$ is a t-norm, that is, a commutative, associative, non-decreasing operation on $[0,1]$, with 1 being the neutral element ($1 * x = x * 1 = x, \forall x \in [0,1]$) and 0 being the absorbent element ($0 * x = x * 0 = 0, \forall x \in [0,1]$). Noticeable t-norms are min, product, and Lukasiewicz operation ($l(x,y) = \max\{0, x+y-1\}$).

Note that the complement of the probabilistic distance, defined as $s(\prec_1, \prec_2) = 1 - \delta_P(\prec_1, \prec_2)$ – the probability that two users with preference orders $\prec_1$ and $\prec_2$ will have the *same* preference over a uniformly randomly chosen pair $(a,b)$ of alternatives – is a fuzzy similarity relation with respect to Lukasievicz t-norm.

## 3 DISTANCE MEASURES ON PARTIALLY SPECIFIED PREFERENCE ORDERS

While the distance functions proposed in the previous section provide various similarity measures that could be used to cluster preferences of multiple users, they are not much good for preference elicitation. For the purpose of elicitation we need to be able to compute the distance when at least one of the preference orders is only *partially* specified.

We first clarify what we mean by "partially specified" preference orders. Recall that a completely specified preference structure $\prec$ on the set $M$ of alternatives is a weak order, i.e., an asymmetric, negatively transitive binary relation on $M$. For a decision maker whose preference structure we have completely elicited, given any two alternatives $a$ and $b$, we know that there are only three possibilities: either she prefers $b$ to $a$ ($a \prec b$), $a$ to $b$ ($b \prec a$), or indifferent between $a$ and $b$ ($a \sim b$). However, when we have little information about the preferences of the decision maker, it might be the case that we can not say anything about her preferences over some two alternatives $a$ and $b$, meaning that none of the above three possibilities applies. In such cases, we say that the preference between $a$ and $b$ is not specified, or $a$ and $b$ are *incomparable*, denoted by $a \parallel b$. Thus, a partially specified order can be viewed as a *partial order*, or *generalized weak order* where for any pair of alternatives $(a,b)$, exactly one of the four relations $\prec, \succ, \sim$, and $\parallel$ holds.

For simplicity, we will call a partially specified preference order, which is a 3-tuple $(\prec, \sim, \parallel)$, a *partial preference order* and denote it by $\prec$. We need a slightly different definition of consistent functions – functions that are consistent with partial preference orders.

**Definition 2** *A real-valued function $f : M \to \Re$ is said to be* consistent *with a partial preference order $\prec$ on $M$ if for all $a,b \in M$, $a \prec b \Rightarrow f(a) < f(b)$ and $a \sim b \Rightarrow f(a) = f(b)$.*

Note that the above definition does not specify what happens when $a \parallel b$. Intuitively, consistent functions capture all information contained in the partial orders, and they might contain more than that.

We now turn our attention to defining a distance measure among partially specified preference orders. In this paper we focus on the certainty case and define a distance measure among partial orders on $\Omega$.

Suppose the preference structures of two users are partially specified by two partial preference orders $(\Omega, \prec_i), i = 1, 2$. Let $V_i^*$ be the set of all functions that are consistent with partial order $\prec_i$, and $V_i$ be the set of equivalence classes of $V_i^*$ with respect to the strategic equivalence of *value* functions. We can also view $V_i$ as a subset of $V_i^*$ that contains strategically different functions, and any member function of $V_i^*$ is strategically equivalent to some member of $V_i$. We will use this interpretation from now on.

We can also view each partial order $\prec_i$ as a set of complete orders that are consistent with it, where consistency means that any relation between any pair of elements that holds with respect to the partial order also holds with respect to the complete orders. In the literature of graphs and orders, partial orders are also called *posets*, and the consistent complete orders of a poset are called its *linear extensions*. It is not hard to see that $\{\prec_{v_i}: v_i \in V_i\}$ is just the set of linear extensions of the partial order $\prec_i$. For simplicity, we will refer to $v_i$ as a linear extension, and $V_i$ is the set of all linear extensions of $\prec_i$.

So the problem now reduces to the problem of defining a distance measure on the space of finite posets over $\Omega$. To the best of our knowledge, there is no general theory that addresses this problem. In this paper we study two approaches to defining such a measure.

**Average-Case Distance**

In the first approach, we consider the average-case behavior of the partial orders. We define the distance between two partial orders $\prec_1$ and $\prec_2$ to be the average of the distances between pairs of complete orders that are consistent with $\prec_1$ and $\prec_2$, respectively. The distance between two complete orders can be any of the distance measures discussed in the previous section: Euclidean ($\delta_E$), Spearman's footrule ($\delta_S$), and probabilistic ($\delta_P$). Formally:

$$\delta_X(\prec_1, \prec_2) = \text{avg}_{(v_1,v_2) \in V_1 \times V_2} \{\delta_X(v_1, v_2)\},$$

where $X$ can be any of the letters "$E$", "$S$", and "$P$" that denote Euclidean distance, Spearman's footrule, and probabilistic distance, respectively. Note that these measures are *not* metrics on the set of partial orders, since the distance between two identical partial orders that are not complete orders is always positive. But this is desirable if the two orders represent the preferences of two different users, since the complete preference orders for the two may actually differ.

The next question is how to compute these distances. A simplistic answer is to compute $\delta_X(v_1, v_2)$ for all pairs of linear extensions $(v_1, v_2) \in V_1 \times V_2$, which involves generating all linear extensions of the partial orders. This approach is computationally prohibitive since the number of linear extensions of a poset can be exponential in terms of its cardinality. In fact, the much easier problem of *counting linear extensions* of posets, a fundamental problem in the theory of ordered sets with applications in computer science (sorting) and social sciences, was shown to be #P-complete [3] by Brightwell and Winkler [1].

Given the hardness of counting and generating linear extensions, we turn to approximation techniques to estimate $\delta_X(\prec_1, \prec_2)$. Let $Y$ be the random variable defined as $Y = \delta_X(v_1, v_2)$, where $v_1, v_2$ are independent uniform random variables on $V_1$ and $V_2$ respectively. We note that $E\,Y = \delta_X(\prec_1, \prec_2)$. We thus can appeal to the Monte Carlo simulation method to estimate $\delta_X(\prec_1, \prec_2)$, provided that we have an efficient algorithm to generate $v_i$ uniformly randomly from $V_i$.

It turns out that counting (approximately) and generating (uniformly randomly) elements of large combinatorial sets are two closely related problems. In fact, Sinclair [15] showed that an efficient algorithm for one problem can be used to construct an efficient algorithm for the other, provided the combinatorial sets have a certain structural property called *self-reducibility*. The set of linear extensions of a poset has this property and, not suprisingly, a number of algorithms for generating (almost) uniformly randomly linear extensions of posets have been developed [8, 2] in order to address the fundamental problem of counting linear extensions. These algorithms are all randomized algorithms based on the Markov chain Monte Carlo technique [4]. In the Appendix we describe the best known algorithm, due to Bubley and Dyer [2] that has a run-ning time of $O(n^3 \log n\epsilon^{-1})$, where $n$ is the poset's cardinality, and $\epsilon$ is the desired accuracy.

---

[3]The complexity class #P, introduced by Valiant [18], consists of all counting problems whose solutions are the number of accepting states of some non-deterministic polynomial-time Turing Machine. A counting problem is *#P-complete* if the problem of counting the number of satisfying assignments to a 3-SAT problem can be reduced to it in polynomial time. #P-complete problems, which are analog counting counterparts of NP-complete problems, are considered very difficult, especially in the view of Toda's results [17], which implies that one call to a #P-complete oracle suffices to solve any problem in the polynomial hierarchy in deterministic polynomial time.

[4]See [7] for a recent survey of this method.

Now with the help of the routine that almost uniformly randomly generates linear extensions of a poset, we can estimate $\delta_X(\prec_1, \prec_2)$ by randomly generating $v_{ij} \in V_i (i = 1, 2; j = 1, \ldots, k)$, computing $\delta_X(v_{1j}, v_{2j}), j = 1, \ldots, k$, and taking the sample mean $\hat{\delta}_X = \frac{1}{k} \sum_{j=1}^{k} \delta_X(v_{1j}, v_{2j})$. This sample mean is an unbiased estimator [5] of $E\,Y = \delta_X(\prec_1, \prec_2)$ with variance $(\text{Var}\,Y)/k$. We can derive a confidence interval for $\delta_X$ as follows. Let $t$ be the ratio of $Y$'s variance and square of its expectation: $t = \text{Var}\,Y / (E\,Y)^2$, a non-negative quantity that can usually be bounded above by $\tau$, which is polynomial in terms of $n$, the input size. Thus $\text{Var}\,Y \leq \tau (E\,Y)^2$ and $\text{Var}\,\hat{\delta}_X \leq (\tau (E\,Y)^2)/k = (\tau \delta_X^2)/k$. For any positive number $c$, Chebysev's inequality says that:

$$\Pr((\hat{\delta}_X - \delta_X)^2 > c\,\text{Var}\,\hat{\delta}_X) \leq 1/c,$$

and thus:

$$\Pr((\hat{\delta}_X - \delta_X)^2 > c\tau \delta_X^2 / k) \leq 1/c,$$

or equivalently:

$$\Pr((1 - \sqrt{c\tau/k})\delta_X \leq \hat{\delta}_X \leq (1 + \sqrt{c\tau/k})\delta_X) \geq 1 - 1/c.$$

As a consequence, if we want our estimator $\hat{\delta}_X$ to be within a multiplicative factor of $1 + \epsilon$ of $\delta$ with probability of at least $1 - 1/c$, it is sufficient to take a sample of size $k = \lceil 4c\tau/\epsilon^2 \rceil$.

There is also another possible way to define the distance measure among partial orders based on their average-case behavior. Note that in the previous section, the Euclidean distance and the Spearmans's footrule are defined among complete orders based on the heights of the outcomes. Accordingly, we can define the corresponding, *generalized Euclidean distance* and *generalized Spearman' footrule* based on the *average* heights of the outcomes with respect to their consistent complete orders. In the literature on ordered sets, the average height of an element with respect to a partial order is simply called its *height*. Denote the height of an element $s_j, j = 1, \ldots, n$ with respect to the partial order $\prec_i, i = 1, 2$ by $h_{ij}$, we can define the distance between $\prec_1$ and $\prec_2$ as:

**Generalized $\delta_S$** :   $\delta'_S(\prec_1, \prec_2) := \frac{1}{2} \sum_{j=1}^{n} |h_{1j} - h_{2j}|$

**Generalized $\delta_E$** :   $\delta'_E(\prec_1, \prec_2) = \sqrt{\sum_{j=1}^{n} (h_{1j} - h_{2j})^2}$.

---

[5]To be more precise, $\hat{\delta}_X$ is *not* an unbiased estimator for $\delta_X$, since the routine only generates *almost* uniform linear extensions. The incurred bias is insignificant and often simply ignored in Markov chain Monte Carlo analysis.

In contrast to the previously discussed average-case distance measures, these distance measures are metrics on the set of partial orders over $S$. Determining the heights of elements of a poset, however, is also a #P-complete problem [1] and requires the approximation technique described above.

**Extreme-Case Distance**

In the second approach, we consider the extreme behaviors of the partial orders $\prec_i$. Specifically, we define the distance $\Delta_X(\prec_1, \prec_2)$ to be an *interval* whose endpoints are the minimum and the maximum of $\delta_X(\prec_{v_1}, \prec_{v_2})$, where $v_i \in V_i, i = 1, 2$, and $X$ is either "$E$", or "$S$", or "$P$":

$$\Delta_X(\prec_1, \prec_2) = [\min_{v_i \in V_i} \delta_X(\prec_{v_1}, \prec_{v_2})\ \max_{v_i \in V_i} \delta_X(\prec_{v_1}, \prec_{v_2})].$$

This approach gives us the flexibility in defining the concept of closeness among partial preference orders. For example, given three partial preference orders $\prec_i, i = 1, 2, 3$, we can take the conservative approach and say that $\prec_1$ is closer to $\prec_2$ than to $\prec_3$ if the upper bound of $\Delta_X(\prec_{v_1}, \prec_{v_2})$ is less than the lower bound of $\Delta_X(\prec_{v_1}, \prec_{v_3})$. We can also take other approaches such as the optimistic approach (*minimin*: closer when the lower bound is smaller) and pessimistic approach (*minimax*: closer when the upper bound is smaller).

Computing $\Delta_X(\prec_1, \prec_2)$ seems to be a difficult combinatorial optimization problem. A reasonable approach is to take the minimum and the maximum of a random sample of sufficiently large size as approximations for the bounds of $\Delta_X$.

## 4 AN IMPLEMENTATION

In this section, we describe MovieFinder [6], an experimental recommender system that recommends movies for people based on this case-based preference elicitation approach. Our main goal of this experiment is to see if the probabilistic distance $\delta_P$ is a reasonble measure of distance on preference structures.

The problem of deciding what movie to see is one of certainty, where the outcomes are the movies themselves. Each movie is characterized by 5 attributes: director, casting, genre, star rating, and time length. We interviewed 10 graduate students to fully elicit their preferences over a set of 50 movies. We then stored these preference structures in our case base in the form of 10 value functions. We then chose one student, say Xaviera, out of the 10 students and "simulated" a repetition of the elicitation of her preferences. We looked

---

[6]The system is available on the World Wide Web at http://cs.uwm.edu/ nguyen/movie/moviemain.htm

at the partial preference structure of Xaviera at various points during the elicitation process and estimated its probabilistic distance from the 10 preference structures in the case base (one of which is hers), using the method described in Section 3.

At the beginning, there was no information about Xaviera's preferences, i.e., the partial preference structure we had was vacuous. This structure was of an equal distance from the structures in the case base, with large variances. We expected that as the simulation progressed, the partial preference structure would become more informative, and the set of closest matches in the case base would become increasingly smaller, eventually to a set of preference structures that were strategically equivalent with Xaviera's preference structure. Our experiments with various students confirmed this expectation.

In order to quickly narrow down the set of closest matches, it was crucial to ask the elicitation questions in the right order. For example, if half of the 50 films are action films and the other half are drama films, and half of the 10 students prefer action films to drama films, then the answer to a single question about preferences over the attribute "genre" can give us preferences over $25 \times 25 = 625$ pairs of movies, which in turn can halve the size of the set of closest matches.

## 5   DISCUSSION

Choosing an appropriate distance measure is only the first step in realizing our envisioned case-based approach to preference elicitation and several difficult technical problems remain to be solved. The first is how to represent the case base. Since our analyses suggest that computing the distance between two preference structures can be computationally complex, it may be desirable to reduce the number of structures with which we must perform comparisons. One possibility is to perform hierarchical clustering on the case base and to store prototype representations of each cluster. We could then use the hierarchical organization to guide the search for a best matching case or could retrieve one of the prototypes. Our distance measure could be used for the clustering but how to create a prototype representation of a set of preferences is an open question.

In the case of certainty, another possible way to speed up the process of identifying the closest match is to limit the number of outcomes and thus to reduce the complexity of distance computing algorithms. For example, suppose that there are 100 outcomes, and we take into consideration only the top 10 outcomes with respect to each partial order. In other words, we ignore information regarding the suboptimal outcomes in determining the distance among partial preference structures. For each pair of partial orders, we then compute their distance based on their restrictions to the union of their corresponding top 10 elements, which is a set of at most 20 elements. This approach computes only an approximate of the actual distance, but can provide significant computational savings.

In a recent paper, Chajewska et al. [3] discuss an approach to preference elicitation similar to ours. Given a data base of user utility functions, they propose clustering them and describing each cluster by a prototype. They propose building a decision tree for associating a user with a prototype utility function based on some elicited pairwise preferences. Their approach requires having a data base of complete utility functions. Their retrieval scheme depends on asking the user questions and ruling out utility functions that conflict with the user's answers. Since no prototype is likely to exactly match the user's preferences, this approach has the problem that the utility function retrieved is sensitive to the order in which questions are asked. In contrast, our approach would retrieve the closest matching preference structure, independent of the order of questions. Since Chajewska and Getoor are initially focusing on the problem of building a working system and we are initially focusing on the theoretical underpinnings, we see their work as complementary to ours.

The closest matching preference structure will in most cases not perfectly match all the preferences expressed by the user. In this case, we would want to modify it to incorporate the unrepresented preferences. For preferences that are simply lacking in the stored structure, this is trivial. For preferences that conflict, we would want to reconcile these differences by modifying the retrieved structure in some minimal way. Minimality could be defined relative to our distance measure.

The case-based approach we are advocating was inspired by the work on collaborative filtering [13, 9], in which the filtering system predicts how interesting a user will find items he has not seen based on the ratings that other users give to items. Each user in a population rates various alternatives, e.g. newsgroup postings or movies, according to a numeric scale. The system then correlates the ratings in order to determine which users' ratings are most similar to each other. Finally, it predicts how well users will like new articles based on ratings from similar users. The work on collaborative filtering is not cast in the framework of decision theory and no theoretical framework or justification are provided for the similarity measures used. The work that we present here can be viewed as an attempt to provide a formal basis for some of the work in this area.

Acknowledgements

The authors would like to thank Hien Nguyen for the implementation of the MovieFinder system, and Tri Le for many helpful discussions. This work was partially supported by NSF grant IRI-9509165.

Appendix

We describe below an algorithm, due to Bubley and Dyer [2], that almost uniformly randomly generates linear extensions of a partial order. The algorithm has running time of $O(n^3 \log n\epsilon^{-1})$, where $n$ is the number of the elements of the partial order, and $\epsilon$ is the desired accuracy, which means that the generated random linear extension has a probability distribution that is within a total variation distance [7] of $\epsilon$ from the uniform distribution. The running time required to obtain a certain precision $\epsilon$ is often called the *mixing time* of the Markov chain. A Markov chain with a mixing time polynomial with respect to the input size (which is the number of elements of the partial order in this case) and $\epsilon^{-1}$ is called *rapidly mixing*.

Suppose that the partial order $\prec$ has $n$ elements, and $N = \{1, 2, \ldots, n\}$. We encode the orderings of these elements with the permutations of the elements of $N$, and the set of linear extensions of $\prec$ by a subset $\mathcal{LE}(\prec)$ of the set of all permutations of the elements of $N$.

For a given concave probability distribution $f$ on $\{1, 2, \ldots, n-1\}$, define a Markov chain $\mathcal{M}_f = \{S_t\}_{t \geq 0}$ on $\mathcal{LE}(\prec)$ as follows. At any time point $t \geq 0$, toss a fair coin. If the coin lands head, then let $S_{t+1} = S_t$. If the coin lands tail, then choose an index $i \in \{1, 2, \ldots, n-1\}$ according to the distribution $f$. If the permutation obtained from $S_t$ by switching the $i$-th and $(i+1)$-st elements of $S_t$ is also a linear extension of $\prec$, i.e., an element of $\mathcal{LE}(\prec)$, then let $S_{t+1}$ be this new permutation. Otherwise, let $S_{t+1} = S_t$.

It is easily seen that $\mathcal{M}_f$ is ergodic with uniform stationary distribution. When $f$ is the uniform distribution on $\{1, 2, \ldots, n-1\}$, $\mathcal{M}_f$ is the Karzanov-Kachiyan chain with mixing time $O(n^5 \log n + n^4 \log \epsilon^{-1})$ [8]. Bubley and Dyer showed that if $f$ is defined as $f(i) = i(n-i)/K$, where $K = (n^3 - n)/6$, then $\mathcal{M}_f$ has mixing time of $O(n^3 \log n\epsilon^{-1})$.

---

[7]The total variation distance between two discrete distributions $P, Q$ over a finite sample space $S$ resembles the Spearman's footrule, and is defined as $d_{TV}(P, Q) = \frac{1}{2}\sum_{s \in S}|P(s) - Q(s)|$.